\documentclass[sigconf]{acmart}

\setcopyright{none}

\AtBeginDocument{%
  }

\copyrightyear{2025}
\acmYear{2025}
\setcopyright{acmlicensed}\acmConference[WWW Companion '25]{Companion Proceedings of the ACM Web Conference 2025}{April 28-May 2, 2025}{Sydney, NSW, Australia}
\acmBooktitle{Companion Proceedings of the ACM Web Conference 2025 (WWW Companion '25), April 28-May 2, 2025, Sydney, NSW, Australia}
\acmDOI{10.1145/3701716.3735080}
\acmISBN{979-8-4007-1331-6/2025/04}

\usepackage{graphicx}
\usepackage{algorithm}
\usepackage{algorithmic}
\usepackage{multirow}
\usepackage{tabularx}
\usepackage{url}
\usepackage{graphicx}
\usepackage{booktabs}
\usepackage{subfig}
\usepackage{subcaption}
\usepackage{csquotes}

\begin{document}

\title{EmoMeta: A Multimodal Dataset for Fine-grained Emotion Classification in Chinese Metaphors}

\author{Xingyuan Lu}
\affiliation{%
\department{School of Software}
  \institution{Dalian University of Technology}
   \city{Dalian}
  \country{China}}
\email{775182652@qq.com}

\author{Yuxi Liu}
\affiliation{%
\department{School of Software}
  \institution{Dalian University of Technology}
   \city{Dalian}
  \country{China}}
\email{liuyx@mail.dlut.edu.cn}

\author{Dongyu Zhang}
\authornote{Corresponding author}
\affiliation{%
\department{School of Foreign Languages and School of Software}
  \institution{Dalian University of Technology}
   \city{Dalian}
  \country{China}}
% \additionalaffiliation{%
%   \department{School of Foreign Languages}
%   \institution{Dalian University of Technology}
%   \city{Dalian}
%   \country{China}
% }
\email{zhangdongyu@dlut.edu.cn}

\author{Zhiyao Wu}
\affiliation{%
\department{Faculty of Business Administration}
  \institution{University of Macau}
  \city{Macau}
  \country{China}
}
\email{bc31072@um.edu.mo}

\author{Jing Ren}
\authornotemark[1]
\affiliation{%
\department{School of Computing Technologies}
  \institution{RMIT University}
  \city{Melbourne}
  \country{Australia}
}
\email{jing.ren@ieee.org}

\author{Feng Xia}
\affiliation{%
\department{School of Computing Technologies}
  \institution{RMIT University}
   \city{Melbourne}
  \country{Australia}}
\email{f.xia@ieee.org}



\begin{abstract}
  Metaphors play a pivotal role in expressing emotions, making them crucial for emotional intelligence. The advent of multimodal data and widespread communication has led to a proliferation of multimodal metaphors, amplifying the complexity of emotion classification compared to single-mode scenarios. However, the scarcity of research on constructing multimodal metaphorical fine-grained emotion datasets hampers progress in this domain. Moreover, existing studies predominantly focus on English, overlooking potential variations in emotional nuances across languages. To address these gaps, we introduce a multimodal dataset in Chinese comprising 5,000 text-image pairs of metaphorical advertisements. Each entry is meticulously annotated for metaphor occurrence, domain relations and fine-grained emotion classification encompassing joy, love, trust, fear, sadness, disgust, anger, surprise, anticipation, and neutral. Our dataset is publicly accessible\footnote{\url{https://github.com/DUTIR-YSQ/EmoMeta}}, facilitating further advancements in this burgeoning field.
\end{abstract}





\maketitle

\section{Introduction}
Metaphorical expressions are ubiquitous in human communication, appearing on average in approximately every third sentence of natural language, as demonstrated by empirical studies. According to \citet{lakoff1980metaphorical}, metaphors allow humans to use a concrete or familiar concept (source domain) to understand an abstract or complex concept (target domain), enhancing both reasoning and communication. For example, in the metaphorical expression \enquote{time is money}, the concept of \enquote{money} is used to underscore the value of \enquote{time}. Metaphor involves a mapping process in which a target domain is conceptualized or understood through the lens of a source domain.

\begin{figure}[htbp]
    \centering
    \begin{minipage}[t]{0.3\linewidth}
        \centering
        \includegraphics[width=\linewidth, height=3cm]{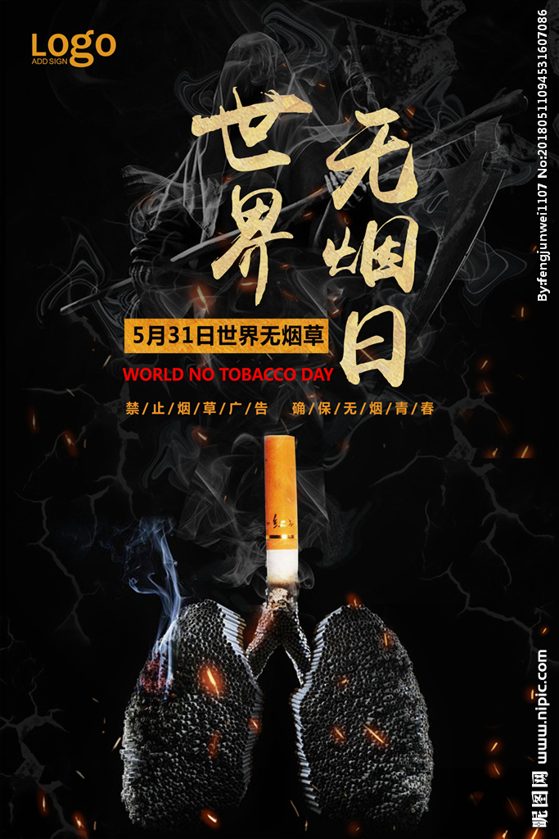}%\\
        \caption*{\footnotesize  (a) World No Tobacco Day, ban tobacco advertising, ensure smoke-free youth.}
    \end{minipage}%
    \hfill%
    \begin{minipage}[t]{0.3\linewidth}
        \centering
        \includegraphics[width=\linewidth, height=3cm]{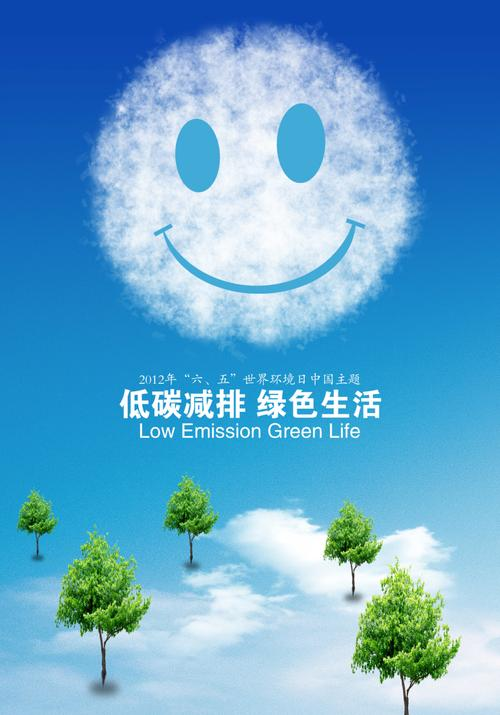}%\\
        \caption*{\footnotesize (b) Low carbon emissions reduction and green living.}
    \end{minipage}%
    \hfill%
    \begin{minipage}[t]{0.3\linewidth}
        \centering
        \includegraphics[width=\linewidth, height=3cm]{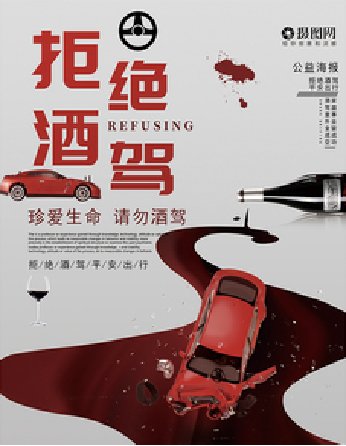}%\\
        \caption*{\footnotesize (c) Refuse drunk driving, cherish life, do not drive under the influence of alcohol.}
    \end{minipage}%
    \vspace{-2mm}    
    \caption{Examples of multimodal metaphors}
    \label{T}

\end{figure}

With the rise of modern media, the prevalence of multimodal metaphors has markedly increased, surpassing their monomodal counterparts due to their vivid, captivating, and persuasive effects. Multimodal metaphors, as defined by \citet{forceville2009multimodal}, involve mapping one domain onto another domain using various modes such as text and image, text and sound, or image and sound. Figure~\ref{T}(a) provides a compelling illustration: a metaphorical representation depicting lungs composed of cigarettes. This visual metaphor symbolically connects two distinct entities—the lung and the cigarette—evoking the idea that smoking is a primary cause of lung damage. The image of the \enquote{cigarette}, representing the source domain, intricately intertwines with the textual and visual representation of the \enquote{lung}, illustrating the impact of smoking on respiratory health.

Emotion, being an abstract and elusive concept, is often conveyed and understood through metaphorical expressions \cite{li2023secret}. Metaphors with emotional connotations play a significant role in implicitly expressing and evoking emotions. For instance, in Figure~\ref{T}(a)\textquotesingle s multimodal metaphor, there is a clear depiction of fear and negativity, whereas Figure~\ref{T}(b)\textquotesingle s metaphor uses both imagery and text to convey joy and positivity. Metaphorical expressions often implicitly and indirectly communicate emotions, contrasting with direct expressions such as \enquote{angry}, \enquote{sad}, or \enquote{happy}. Metaphors are valued for their vividness and ability to evoke mental imagery, making them effective in conveying emotional nuances. 

%\vspace{-1mm}

\vspace{-3mm}

The interaction between emotion and metaphor has been extensively studied by scholars across various disciplines, including psychology \cite{fetterman2016scope, huangfu2025cognitive}, neuroscience \cite{malinowski2015metaphor}, science learning \cite{yonai2025powered}, and natural language processing \cite{mohammad2016metaphor, zhang2024camel}. Notably, with the rapid expansion of mass communication, there has been a considerable increase in multimodal emotional information that includes metaphors. It has become increasingly common for Internet users to employ vivid and colorful multimodal metaphors to express their emotions.

Despite the pivotal role of metaphorical information in advancing emotion computing, the domain of multimodal metaphor-driven emotion computing remains largely unexplored. In natural language processing (NLP), efforts predominantly focus on classifying emotions into basic categories such as positive or negative, often overlooking a more nuanced spectrum of emotional states such as joy or fear. Furthermore, much of this research is conducted in English, which creates a significant gap in our understanding of how metaphors express emotions across different languages. This limitation is particularly notable given the diverse ways in which metaphors can convey emotions across cultures, despite the presence of similar conceptual metaphors in various languages that stem from universal human cognition and physical experiences \cite{kovecses20101metaphor}. The same phrase can carry different meanings across languages. Metaphors vary depending on the linguistic context. For example, the phrase \enquote{This person looks like a dinosaur} might be used in other languages and cultures to imply that a person is very strong, but in Chinese, it is used to describe someone as looking unappealing.  Moreover, some Chinese metaphors may be more subtle and lack direct equivalents in other languages.

To address the gap in data resources mentioned earlier, we introduce our Chinese multimodal fine-grained emotion dataset, comprising 5k metaphorical advertisements with manual annotations for metaphors and labels for 9 emotion categories or Neutral.

\section{Related Datasets}
Current research on metaphorical emotions in typically utilizes broad emotion classifications, such as two-category, three-category or five-category systems. For instance, \citet{mohammad2016metaphor} divide emotions into emotional and not emotional. \citet{zhang2023multicmet} simplify them into negative, neutral and positive, while \citet{zhang2021multimet} categorize emotions into very negative, negative, neutral, positive, and very positive. However, these datasets mainly focus on annotating the polarity and degree of emotions. There remains a relative scarcity of fine-grained emotion classification. To address this gap, inspired by \citet{demszky2020goemotions, pant2023phymer, zhang2024mersa, lim2024integrating}, we have developed a fine-grained emotion classification system tailored for metaphorical emotions in advertisements. The system is based on the six fundamental emotions proposed by \citet{ekman1992argument}( joy, anger, fear, sadness, disgust, and surprise) as well as the additional basic emotion categories suggested by \citet{plutchik1980general}, which include trust and anticipation. The interactions between different emotions can be categorized based on different datasets \cite{zaid2025power}.

\section{EmoMeta}
Our dataset stands out as a multimodal compilation, encompassing both images and corresponding textual content aligned with the images. Specifically, it comprises 5k metaphorical advertisements, meticulously labeled across nine emotion classifications plus a Neutral category. Notably, these advertisements are categorized into two types: public service advertisements and commercial advertisements. The rationale behind selecting advertisements as our primary data source stems from their inherent characteristics. Advertisements, by nature, exhibit distinct emotional tendencies, effectively conveying emotions through succinct text and compelling imagery. This makes advertisements particularly well-suited for emotion classification tasks. Furthermore, the concise text and vivid imagery prevalent in advertisements render them valuable for various other language-related tasks. It is worth noting that several datasets composed of advertisements exist to offer robust data support for natural language processing tasks, exemplified by \citet{forceville2017visual} and the work by \citet{zhang2021multimet}.

\subsection{Data Collection and Process}
Our dataset draws from four distinct data sources. The primary source involves extracting advertisements with both images and text from a previously established dataset \cite{zhang2023multicmet}. Secondly, we curate potential Chinese metaphorical advertising samples from a comprehensive commercial advertising dataset. This dataset incorporates images and internal texts sourced from the IFlytek Advertising Image Classification Competition, released in 2021\footnote{\url{https://aistudio.baidu.com/aistudio/datasetdetail/102279}}, serving as the second foundational data source. However, this part of the data is not accompanied by a URL in the dataset of the data source. Therefore, we upload the images in the form of files, following the original data form of the data source. We strictly adhere to the terms and conditions governing the use of these publicly available datasets. The third source involves a meticulous search across prominent search engines, specifically Baidu and Bing, which are widely popular and possess more extensive resources. We have adhered to the data use agreements of these search engine platforms.
After collecting the data, we validated image uniqueness using MD5 encoding \cite{rivest1992md5}. We manually curated the dataset by excluding blurry images, those not categorized as advertisements, and ads without corresponding text. To extract text from images, we used the Paddle OCR model \cite{du2020pp}, but this sometimes resulted in errors like typos, strange symbols, and word order issues. We manually corrected these by cross-referencing the text with the images. We also removed non-Chinese text, especially English, and excluded irrelevant elements like publishers and number strings to ensure a clean dataset for analysis.

\subsection{Annotation Model}
\begin{figure}[htbp]
  \centering
  \includegraphics[width=0.35\textwidth]{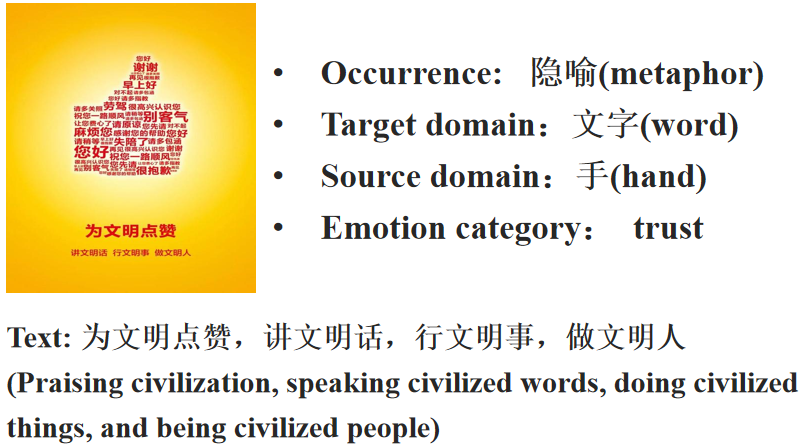}
  \vspace{-2mm} 
  \caption{An example of a metaphorical annotation}
  \label{Ansam}
\end{figure}

We meticulously annotated the text-image pairs based on the presence of metaphors, distinguishing between literal and metaphorical instances. In cases where all data is metaphorical, we specifically note this condition. Our annotations also include identification of the source domain and target domain, considering target/source words in the text, verbalized target/source words in the image, or a combination of target/source words in both the text and image. Additionally, we annotated the emotion category conveyed by the emotion metaphors. The annotation model is represented as AnnotationModel = (Occurrence, source domain, target domain, emotion category). For clarity, Figure~\ref{Ansam} serves as an example illustrating the annotation. Notably, during the emotion annotation process, if an image conveys no emotion while the text does, the annotation reflects the emotion expressed in the text. Conversely, if the image conveys emotion but the text does not, the annotation reflects the emotion in the image. When there is a conflict between the emotions in the image and the text, this often signifies a particularly strong metaphorical emotion. In such cases, a comprehensive evaluation is necessary, with the more intense emotion from either source being used as the final annotation.

%\begin{figure}[!t]

\vspace{-2mm} 

\subsection{Annotation}
\textbf{Metaphorcial or literal.}
Building on the methodologies of \citet{zhang2021multimet} and \citet{zhang2023multicmet}, our approach to determining if a sample is metaphorical focuses on the relational level, identifying metaphorical links between the source and target domains. This judgment involves both verbal and visual aspects, as detailed by \citet{phillips2004beyond}. For text, annotations are based on language and context, while for images, they depend on visual features. Annotators evaluate the metaphorical nature of samples with both text and images by examining inconsistencies carefully. They evaluate and explain an irreversible \enquote{A is B} identity relationship based on the textual or visual elements present in the sample. This comprehensive approach ensures a thorough assessment of metaphorical content across both modalities.

\noindent\textbf{Target domain and source domain.}
Metaphor, as described by \citet{lakoff1980metaphorical}, involves mapping a concrete concept from one domain to an abstract concept in another. People understand unfamiliar elements in the abstract domain by comparing them with familiar elements in the concrete domain. Here, the abstract domain represents the target domain, while the concrete domain represents the source domain. In essence, the target domain is the concept conveyed through the metaphor, and the source domain is the concept used to express it. The effectiveness of a metaphor depends on the similarity between the source and target domains. For instance, consider the metaphorical statement \textquotesingle time is money\textquotesingle . In this case, \textquotesingle time\textquotesingle (target domain) represents the abstract concept in the abstract domain, while \textquotesingle money\textquotesingle (source domain) embodies the concrete concept in the concrete domain. The metaphor works due to the perceived similarity between time and money, both being regarded as precious resources. Additionally, the sentence conveys the treasure of time.

\noindent\textbf{Emotion categories.}
\citet{ekman1992argument} originally proposed a set of six basic emotions comprising \textquotesingle joy\textquotesingle , \textquotesingle sadness\textquotesingle , \textquotesingle anger\textquotesingle , \textquotesingle fear\textquotesingle , \textquotesingle disgust\textquotesingle , and \textquotesingle surprise\textquotesingle . In contrast, \citet{plutchik1980general} expanded upon Ekman\textquotesingle s model by including the six basic emotions and introducing \textquotesingle trust\textquotesingle  and \textquotesingle anticipation\textquotesingle . However, these eight emotional classifications do not align entirely with the contemporary definitions of these emotions. Both Ekman\textquotesingle s and Plutchik\textquotesingle s theories give rise to numerous other emotions derived from these foundational eight, and the interplay of emotions, involving deepening, reduction, and fusion, can generate novel emotional states. 
It\textquotesingle s important to note that the classification of emotions can vary significantly due to different sources of samples \cite{mohammad2021sentiment}. We tailor our emotional classification for advertising. In particular, some advertisements may not overtly exhibit emotional expressions. Therefore, we classify emotions for advertising into \textquotesingle joy\textquotesingle , \textquotesingle love\textquotesingle , \textquotesingle trust\textquotesingle , \textquotesingle fear\textquotesingle , \textquotesingle sadness\textquotesingle , \textquotesingle disgust\textquotesingle , \textquotesingle anger\textquotesingle , \textquotesingle surprise\textquotesingle , \textquotesingle anticipation\textquotesingle , and \textquotesingle neutral\textquotesingle, as follow.

Joy encompasses a sense of happiness, optimism, and relaxation, embracing feelings of tranquility and ecstasy. Love is a profound and positive emotional and psychological state, signifying deep and sincere affection towards individuals or entities. Trust involves the belief that someone or something is good, sincere, and honest. It encompasses emotions associated with acceptance, liking, and appreciation. Fear conveys a negative sensation that arises in the face of danger or when confronted with something frightening. Sadness is commonly employed to characterize the psychological state experienced when confronting negative emotions like loss and pain. Disgust denotes a profound aversion towards someone or something deemed unacceptable, distasteful, or possessing unpleasant visual or olfactory qualities. Anger is a potent emotion that surfaces when confronted with something bad or unjust. It encompasses feelings of trouble and rage, including annoyance and intense displeasure. Surprise is the emotion elicited by unforeseen or sudden events, manifesting in a state of distraction and amazement. Anticipation conveys a sense of excitement about an impending event, typically perceived as favorable.

\subsection{Annotation Process and Quality Control}
We enlisted native Chinese-speaking researchers with a background in metaphor research as annotators, organized into three groups of three. The first two groups worked independently, and if disagreements arose, the third group re-evaluated the annotations. After discussions, a final decision was made, and when consensus couldn’t be reached, a majority vote determined the outcome. We used the Kappa score \cite{fleiss1971measuring} to assess consistency, with scores of $\alpha$ = 0.68, $\alpha$ = 0.61, $\alpha$ = 0.63, and $\alpha$ = 0.58 for metaphor identification, source domain, target domain, and emotion categories, showing reliable annotations. We also considered source and target domains consistent if they differ in wording but convey the same meaning.

\vspace{-1mm} 

\begin{figure}[ht]
  \centering
  \includegraphics[width=0.46\textwidth]{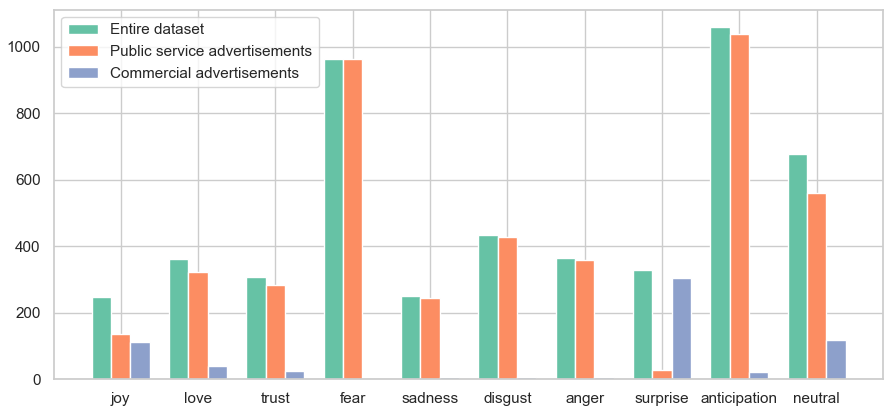}
  \vspace{-2mm} 
  \caption{Emotional distribution of EmoMeta}
  \label{dataw}
\end{figure}

\vspace{-2mm} 

\section{Dataset Analysis}

The proportion of public service advertisements and commercial advertisements is 87\% and 13\% respectively. The detailed statistical analysis of emotion classifications is shown in Figure~\ref{dataw}. It reveals that \textquotesingle fear\textquotesingle and \textquotesingle anticipation\textquotesingle are the most common emotions in the dataset. Additionally, separate analyses for public service and commercial advertisements are also presented in Figure~\ref{dataw}. In the context of public service advertisements, the prevalence of \textquotesingle fear\textquotesingle  and \textquotesingle anticipation\textquotesingle  emotions is notable, while \textquotesingle surprise\textquotesingle  is infrequently observed. This suggests a tendency to convey intense negative emotions, such as \textquotesingle fear\textquotesingle , in metaphorical public service advertisements, aligning with the purpose of creating fear as a warning. Regarding \textquotesingle anticipation\textquotesingle , public service advertisements often signify an expectation for a positive outcome, encouraging others to move in a favorable direction. The emotion of \textquotesingle surprise\textquotesingle  is less reflected in public service advertisements. Analyzing the emotional distribution of commercial advertisements reveals a higher proportion of \textquotesingle surprise\textquotesingle , with \textquotesingle fear\textquotesingle , \textquotesingle sadness\textquotesingle , \textquotesingle disgust\textquotesingle , and \textquotesingle anger\textquotesingle  appearing less frequently. This aligns with the trend in product advertisements, where the expression of negative emotions is less common. \textquotesingle Surprise\textquotesingle  in commercial advertisements often denotes favorable conditions for consumers, such as attractive prices and high-quality materials. From the separate analysis of emotions in public service and commercial advertisements, it is evident that there is a certain degree of complementarity between the two. This complementarity contributes to a more balanced emotional distribution across the entire dataset.

\section{Conclusion}

In conclusion, this paper introduces a pioneering multimodal metaphorical fine-grained emotion dataset in Chinese, annotated manually for metaphoric presence, source domain, target domain, and fine-grained emotion category to address the lack of dataset in the area.  We anticipate that our dataset will contribute valuable data for advancing research on metaphors and emotions.

\bibliographystyle{ACM-Reference-Format}
\bibliography{sample-base}


\begin{thebibliography}{25}

%%% the Web address.

\ifx \showCODEN    \undefined \def \showCODEN     #1{\unskip}     \fi
\ifx \showDOI      \undefined \def \showDOI       #1{#1}\fi
\ifx \showISBNx    \undefined \def \showISBNx     #1{\unskip}     \fi
\ifx \showISBNxiii \undefined \def \showISBNxiii  #1{\unskip}     \fi
\ifx \showISSN     \undefined \def \showISSN      #1{\unskip}     \fi
\ifx \showLCCN     \undefined \def \showLCCN      #1{\unskip}     \fi
\ifx \shownote     \undefined \def \shownote      #1{#1}          \fi
\ifx \showarticletitle \undefined \def \showarticletitle #1{#1}   \fi
\ifx \showURL      \undefined \def \showURL       {\relax}        \fi
% invisible to TeX
\providecommand\bibfield[2]{#2}
\providecommand\bibinfo[2]{#2}
\providecommand\natexlab[1]{#1}
\providecommand\showeprint[2][]{arXiv:#2}

\bibitem[Demszky et~al\mbox{.}(2020)]%
        {demszky2020goemotions}
\bibfield{author}{\bibinfo{person}{Dorottya Demszky}, \bibinfo{person}{Dana Movshovitz-Attias}, \bibinfo{person}{Jeongwoo Ko}, \bibinfo{person}{Alan Cowen}, \bibinfo{person}{Gaurav Nemade}, {and} \bibinfo{person}{Sujith Ravi}.} \bibinfo{year}{2020}\natexlab{}.
\newblock \showarticletitle{GoEmotions: A dataset of fine-grained emotions}.
\newblock \bibinfo{journal}{\emph{arXiv preprint arXiv:2005.00547}} (\bibinfo{year}{2020}).
\newblock


\bibitem[Du et~al\mbox{.}(2020)]%
        {du2020pp}
\bibfield{author}{\bibinfo{person}{Yuning Du}, \bibinfo{person}{Chenxia Li}, \bibinfo{person}{Ruoyu Guo}, \bibinfo{person}{Xiaoting Yin}, \bibinfo{person}{Weiwei Liu}, \bibinfo{person}{Jun Zhou}, \bibinfo{person}{Yifan Bai}, \bibinfo{person}{Zilin Yu}, \bibinfo{person}{Yehua Yang}, \bibinfo{person}{Qingqing Dang}, {et~al\mbox{.}}} \bibinfo{year}{2020}\natexlab{}.
\newblock \showarticletitle{Pp-ocr: A practical ultra lightweight ocr system}.
\newblock \bibinfo{journal}{\emph{arXiv preprint arXiv:2009.09941}} (\bibinfo{year}{2020}).
\newblock


\bibitem[Ekman(1992)]%
        {ekman1992argument}
\bibfield{author}{\bibinfo{person}{Paul Ekman}.} \bibinfo{year}{1992}\natexlab{}.
\newblock \showarticletitle{An argument for basic emotions}.
\newblock \bibinfo{journal}{\emph{Cognition \& emotion}} \bibinfo{volume}{6}, \bibinfo{number}{3-4} (\bibinfo{year}{1992}), \bibinfo{pages}{169--200}.
\newblock


\bibitem[Fetterman et~al\mbox{.}(2016)]%
        {fetterman2016scope}
\bibfield{author}{\bibinfo{person}{Adam~K Fetterman}, \bibinfo{person}{Jessica~L Bair}, \bibinfo{person}{Marc Werth}, \bibinfo{person}{Florian Landkammer}, {and} \bibinfo{person}{Michael~D Robinson}.} \bibinfo{year}{2016}\natexlab{}.
\newblock \showarticletitle{The scope and consequences of metaphoric thinking: Using individual differences in metaphor usage to understand how metaphor functions.}
\newblock \bibinfo{journal}{\emph{Journal of Personality and Social Psychology}} \bibinfo{volume}{110}, \bibinfo{number}{3} (\bibinfo{year}{2016}), \bibinfo{pages}{458}.
\newblock


\bibitem[Fleiss(1971)]%
        {fleiss1971measuring}
\bibfield{author}{\bibinfo{person}{Joseph~L Fleiss}.} \bibinfo{year}{1971}\natexlab{}.
\newblock \showarticletitle{Measuring nominal scale agreement among many raters.}
\newblock \bibinfo{journal}{\emph{Psychological bulletin}} \bibinfo{volume}{76}, \bibinfo{number}{5} (\bibinfo{year}{1971}), \bibinfo{pages}{378}.
\newblock


\bibitem[Forceville(2017)]%
        {forceville2017visual}
\bibfield{author}{\bibinfo{person}{Charles Forceville}.} \bibinfo{year}{2017}\natexlab{}.
\newblock \showarticletitle{Visual and multimodal metaphor in advertising: Cultural perspectives.}
\newblock \bibinfo{journal}{\emph{Styles of communication}} \bibinfo{volume}{9}, \bibinfo{number}{2} (\bibinfo{year}{2017}).
\newblock


\bibitem[Forceville et~al\mbox{.}(2009)]%
        {forceville2009multimodal}
\bibfield{author}{\bibinfo{person}{Charles Forceville}, \bibinfo{person}{Eduardo Urios-Aparisi}, {et~al\mbox{.}}} \bibinfo{year}{2009}\natexlab{}.
\newblock \bibinfo{booktitle}{\emph{Multimodal metaphor}}. Vol.~\bibinfo{volume}{11}.
\newblock \bibinfo{publisher}{Mouton de Gruyter Berlin}.
\newblock


\bibitem[Huangfu and Cheng(2025)]%
        {huangfu2025cognitive}
\bibfield{author}{\bibinfo{person}{Baihui Huangfu} {and} \bibinfo{person}{Wenjuan Cheng}.} \bibinfo{year}{2025}\natexlab{}.
\newblock \showarticletitle{Cognitive computing method based on decoding psychological emotional states}.
\newblock \bibinfo{journal}{\emph{International Journal of Cognitive Computing in Engineering}}  \bibinfo{volume}{6} (\bibinfo{year}{2025}), \bibinfo{pages}{32--43}.
\newblock


\bibitem[K{\"o}vecses(2010)]%
        {kovecses20101metaphor}
\bibfield{author}{\bibinfo{person}{Zolt{\'a}n K{\"o}vecses}.} \bibinfo{year}{2010}\natexlab{}.
\newblock \showarticletitle{Metaphor and culture}.
\newblock \bibinfo{journal}{\emph{Acta Universitatis Sapientiae, Philologica}} \bibinfo{volume}{2}, \bibinfo{number}{2} (\bibinfo{year}{2010}), \bibinfo{pages}{197--220}.
\newblock


\bibitem[Lakoff and Johnson(1980)]%
        {lakoff1980metaphorical}
\bibfield{author}{\bibinfo{person}{George Lakoff} {and} \bibinfo{person}{Mark Johnson}.} \bibinfo{year}{1980}\natexlab{}.
\newblock \showarticletitle{The metaphorical structure of the human conceptual system}.
\newblock \bibinfo{journal}{\emph{Cognitive science}} \bibinfo{volume}{4}, \bibinfo{number}{2} (\bibinfo{year}{1980}), \bibinfo{pages}{195--208}.
\newblock


\bibitem[Li et~al\mbox{.}(2023)]%
        {li2023secret}
\bibfield{author}{\bibinfo{person}{Yucheng Li}, \bibinfo{person}{Frank Guerin}, {and} \bibinfo{person}{Chenghua Lin}.} \bibinfo{year}{2023}\natexlab{}.
\newblock \showarticletitle{The secret of metaphor on expressing stronger emotion}.
\newblock \bibinfo{journal}{\emph{arXiv preprint arXiv:2301.13042}} (\bibinfo{year}{2023}).
\newblock


\bibitem[Lim and Cheong(2024)]%
        {lim2024integrating}
\bibfield{author}{\bibinfo{person}{Dongjun Lim} {and} \bibinfo{person}{Yun-Gyung Cheong}.} \bibinfo{year}{2024}\natexlab{}.
\newblock \showarticletitle{Integrating Plutchik’s Theory with Mixture of Experts for Enhancing Emotion Classification}. In \bibinfo{booktitle}{\emph{Proceedings of the 2024 Conference on Empirical Methods in Natural Language Processing}}. \bibinfo{pages}{857--867}.
\newblock


\bibitem[Malinowski and Horton(2015)]%
        {malinowski2015metaphor}
\bibfield{author}{\bibinfo{person}{Josie~E Malinowski} {and} \bibinfo{person}{Caroline~L Horton}.} \bibinfo{year}{2015}\natexlab{}.
\newblock \showarticletitle{Metaphor and hyperassociativity: the imagination mechanisms behind emotion assimilation in sleep and dreaming}.
\newblock \bibinfo{journal}{\emph{Frontiers in psychology}}  \bibinfo{volume}{6} (\bibinfo{year}{2015}), \bibinfo{pages}{150080}.
\newblock


\bibitem[Mohammad et~al\mbox{.}(2016)]%
        {mohammad2016metaphor}
\bibfield{author}{\bibinfo{person}{Saif Mohammad}, \bibinfo{person}{Ekaterina Shutova}, {and} \bibinfo{person}{Peter Turney}.} \bibinfo{year}{2016}\natexlab{}.
\newblock \showarticletitle{Metaphor as a medium for emotion: An empirical study}. In \bibinfo{booktitle}{\emph{Proceedings of the fifth joint conference on lexical and computational semantics}}. \bibinfo{pages}{23--33}.
\newblock


\bibitem[Mohammad(2021)]%
        {mohammad2021sentiment}
\bibfield{author}{\bibinfo{person}{Saif~M Mohammad}.} \bibinfo{year}{2021}\natexlab{}.
\newblock \showarticletitle{Sentiment analysis: Automatically detecting valence, emotions, and other affectual states from text}.
\newblock In \bibinfo{booktitle}{\emph{Emotion measurement}}. \bibinfo{publisher}{Elsevier}, \bibinfo{pages}{323--379}.
\newblock


\bibitem[Pant et~al\mbox{.}(2023)]%
        {pant2023phymer}
\bibfield{author}{\bibinfo{person}{Sudarshan Pant}, \bibinfo{person}{Hyung-Jeong Yang}, \bibinfo{person}{Eunchae Lim}, \bibinfo{person}{Soo-Hyung Kim}, {and} \bibinfo{person}{Seok-Bong Yoo}.} \bibinfo{year}{2023}\natexlab{}.
\newblock \showarticletitle{PhyMER: Physiological Dataset for Multimodal Emotion Recognition with Personality as a Context}.
\newblock \bibinfo{journal}{\emph{IEEE Access}} (\bibinfo{year}{2023}).
\newblock


\bibitem[Phillips and McQuarrie(2004)]%
        {phillips2004beyond}
\bibfield{author}{\bibinfo{person}{Barbara~J Phillips} {and} \bibinfo{person}{Edward~F McQuarrie}.} \bibinfo{year}{2004}\natexlab{}.
\newblock \showarticletitle{Beyond visual metaphor: A new typology of visual rhetoric in advertising}.
\newblock \bibinfo{journal}{\emph{Marketing theory}} \bibinfo{volume}{4}, \bibinfo{number}{1-2} (\bibinfo{year}{2004}), \bibinfo{pages}{113--136}.
\newblock


\bibitem[Plutchik(1980)]%
        {plutchik1980general}
\bibfield{author}{\bibinfo{person}{Robert Plutchik}.} \bibinfo{year}{1980}\natexlab{}.
\newblock \showarticletitle{A general psychoevolutionary theory of emotion}.
\newblock In \bibinfo{booktitle}{\emph{Theories of emotion}}. \bibinfo{publisher}{Elsevier}, \bibinfo{pages}{3--33}.
\newblock


\bibitem[Rivest(1992)]%
        {rivest1992md5}
\bibfield{author}{\bibinfo{person}{Ronald Rivest}.} \bibinfo{year}{1992}\natexlab{}.
\newblock \bibinfo{booktitle}{\emph{The MD5 message-digest algorithm}}.
\newblock \bibinfo{type}{{T}echnical {R}eport}.
\newblock


\bibitem[Yonai and Blonder(2025)]%
        {yonai2025powered}
\bibfield{author}{\bibinfo{person}{Ella Yonai} {and} \bibinfo{person}{Ron Blonder}.} \bibinfo{year}{2025}\natexlab{}.
\newblock \showarticletitle{“Powered by emotions”: Exploring emotion induction in out-of-school authentic science learning}.
\newblock \bibinfo{journal}{\emph{Journal of Research in Science Teaching}} \bibinfo{volume}{62}, \bibinfo{number}{2} (\bibinfo{year}{2025}), \bibinfo{pages}{553--575}.
\newblock


\bibitem[Zaid et~al\mbox{.}(2025)]%
        {zaid2025power}
\bibfield{author}{\bibinfo{person}{Sumaia~Mohammed Zaid}, \bibinfo{person}{Fonny~Dameaty Hutagalung}, \bibinfo{person}{Harris Shah~Bin Abd~Hamid}, {and} \bibinfo{person}{Sahar~Mohammed Taresh}.} \bibinfo{year}{2025}\natexlab{}.
\newblock \showarticletitle{The power of emotion regulation: how managing sadness influences depression and anxiety?}
\newblock \bibinfo{journal}{\emph{BMC psychology}} \bibinfo{volume}{13}, \bibinfo{number}{1} (\bibinfo{year}{2025}), \bibinfo{pages}{1--12}.
\newblock


\bibitem[Zhang et~al\mbox{.}(2023)]%
        {zhang2023multicmet}
\bibfield{author}{\bibinfo{person}{Dongyu Zhang}, \bibinfo{person}{Jingwei Yu}, \bibinfo{person}{Senyuan Jin}, \bibinfo{person}{Liang Yang}, {and} \bibinfo{person}{Hongfei Lin}.} \bibinfo{year}{2023}\natexlab{}.
\newblock \showarticletitle{MultiCMET: A Novel Chinese Benchmark for Understanding Multimodal Metaphor}. In \bibinfo{booktitle}{\emph{Findings of the Association for Computational Linguistics: EMNLP 2023}}. \bibinfo{pages}{6141--6154}.
\newblock


\bibitem[Zhang et~al\mbox{.}(2021)]%
        {zhang2021multimet}
\bibfield{author}{\bibinfo{person}{Dongyu Zhang}, \bibinfo{person}{Minghao Zhang}, \bibinfo{person}{Heting Zhang}, \bibinfo{person}{Liang Yang}, {and} \bibinfo{person}{Hongfei Lin}.} \bibinfo{year}{2021}\natexlab{}.
\newblock \showarticletitle{Multimet: A multimodal dataset for metaphor understanding}. In \bibinfo{booktitle}{\emph{Proceedings of the 59th Annual Meeting of the Association for Computational Linguistics and the 11th International Joint Conference on Natural Language Processing (Volume 1: Long Papers)}}. \bibinfo{pages}{3214--3225}.
\newblock


\bibitem[Zhang et~al\mbox{.}(2024b)]%
        {zhang2024mersa}
\bibfield{author}{\bibinfo{person}{Enshi Zhang}, \bibinfo{person}{Rafael Trujillo}, {and} \bibinfo{person}{Christian Poellabauer}.} \bibinfo{year}{2024}\natexlab{b}.
\newblock \showarticletitle{The MERSA Dataset and a Transformer-Based Approach for Speech Emotion Recognition}. In \bibinfo{booktitle}{\emph{Proceedings of the 62nd Annual Meeting of the Association for Computational Linguistics (Volume 1: Long Papers)}}. \bibinfo{pages}{13960--13970}.
\newblock


\bibitem[Zhang et~al\mbox{.}(2024a)]%
        {zhang2024camel}
\bibfield{author}{\bibinfo{person}{Linhao Zhang}, \bibinfo{person}{Li Jin}, \bibinfo{person}{Guangluan Xu}, \bibinfo{person}{Xiaoyu Li}, \bibinfo{person}{Cai Xu}, \bibinfo{person}{Kaiwen Wei}, \bibinfo{person}{Nayu Liu}, {and} \bibinfo{person}{Haonan Liu}.} \bibinfo{year}{2024}\natexlab{a}.
\newblock \showarticletitle{CAMEL: Capturing Metaphorical Alignment with Context Disentangling for Multimodal Emotion Recognition}. In \bibinfo{booktitle}{\emph{Proceedings of the AAAI Conference on Artificial Intelligence}}, Vol.~\bibinfo{volume}{38}. \bibinfo{pages}{9341--9349}.
\newblock


\end{thebibliography}

\end{document}